\documentclass{article}

\usepackage{microtype}
\usepackage{graphicx}
\usepackage{subcaption}
\usepackage{booktabs} 

\usepackage{hyperref}
\usepackage{setspace}

\usepackage[accepted]{icml2026}

\usepackage{amsmath}
\usepackage{amssymb}
\usepackage{mathtools}
\usepackage{amsthm}
\usepackage{float}

\usepackage[capitalize,noabbrev]{cleveref}

\theoremstyle{plain}

\theoremstyle{definition}

\theoremstyle{remark}

\usepackage[textsize=tiny]{todonotes}

\usepackage{multirow}
\usepackage{arydshln} 
\usepackage{array}     
\usepackage{dashrule}  
\usepackage{makecell}

\icmltitlerunning{SUN: Shared Use of Next-token Prediction for Efficient Multi-LLM Disaggregated Serving}

\begin{document}

\twocolumn[
  \icmltitle{SUN: Shared Use of Next-token Prediction \\ for Efficient Multi-LLM Disaggregated Serving}

  \icmlsetsymbol{equal}{*}

  \begin{icmlauthorlist}
    \icmlauthor{Sunghyeon Woo}{equal,naver}
    \icmlauthor{Ahreum Seo}{equal,naver}
    \icmlauthor{Jaegwang Lee}{naver}
    \icmlauthor{Jaeeun Kil}{naver}
    \icmlauthor{Hanbae Seo}{naver}
    \icmlauthor{Joonghoon Kim}{naver}
    \icmlauthor{Baeseong Park}{naver}
    \icmlauthor{Se Jung Kwon}{naver}
    \icmlauthor{Dongsoo Lee}{naver}
  \end{icmlauthorlist}

    \vspace{2mm}
  {\centering
    {\normalsize NAVER Cloud\par}
    \vspace{1pt}
    {\normalsize \textnormal{\{sunghyeon.woo1, arr.seo\}@navercorp.com}\par}
  }
  \vspace{2.0mm}
  
  \icmlkeywords{Machine Learning, ICML, Disaggregated Serving, Multi-LLM Serving, KV Cache, Weight Quantization}

  \addvspace{1.5mm}
]

\makeatletter
\renewcommand{\icmlaffiliation}[2]{}%
\renewcommand{\icmlcorrespondingauthor}[2]{}%
\makeatother

\printAffiliationsAndNotice{\icmlEqualContribution}




\newcommand{\method}{\textsc{SUN}}

\begin{abstract}
In multi-model LLM serving, decode execution remains inefficient due to model-specific resource partitioning: since cross-model batching is not possible, memory-bound decoding often suffers from severe GPU underutilization, especially under skewed workloads. We propose Shared Use of Next-token Prediction (SUN), the first approach that enables cross-model sharing of decode execution in disaggregated multi-LLM serving. SUN decomposes a decoder-only Transformer into a prefill module and a decode module, and fine-tunes only the task-specific prefill module, enabling a frozen decode module to be shared across models. This design enables a model-agnostic decode routing policy that balances decode requests across shared workers to maximize utilization. Across diverse tasks and model families, SUN achieves accuracy comparable to full fine-tuning while maintaining system throughput with fewer decode workers. In particular, SUN improves throughput per GPU by up to 2.0$\times$ over conventional disaggregation while keeping time-per-output-token (TPOT) within 5\%. SUN inherently enables and facilitates low-bit decoding; with Quantized SUN (QSUN), it achieves a 45\% speedup with comparable accuracy to SUN while preserving the benefits of shared decoding.


\end{abstract}

\section{Introduction}
\label{sec:introduction}

Recent advances in large language models (LLMs) have expanded their adoption beyond general-purpose chatbots, enabling diverse domain-specific models \cite{parlm2-singhal,chatlaw-cui,bloomberg-wu,codellama-rozi} and agentic AI workflows in which multiple specialized models collaborate to solve complex real-world problems \cite{Shen-toollearners,multiagent_finetuning,mixture_of_agents,toolorchestra-su}. In line with this trend, service providers are increasingly facing the challenge of multi-LLM serving, as they move toward deploying and serving tens or even hundreds of specialized LLMs concurrently. For instance, in a typical multi-agent pipeline, a single user query may invoke a router model, multiple domain experts, and a summarizer, each requiring low-latency decode execution.

A fundamental challenge in multi-LLM serving arises at two levels: intra-model interference between the prefill and decode phases, and inter-model isolation caused by model-specific resource partitioning. Prefill is largely compute-bound due to batched matrix multiplications over the prompt, whereas decode is typically memory-bound because each autoregressive step must read and update a growing key--value (KV) cache with relatively little compute per byte~\cite{kwon-vllm,sglnag-zheng,Zhong-distserve}. When prefill and decode are co-located on the same GPUs, their coupled scheduling leads to prefill--decode interference, which inflates tail latency~\cite{Zhong-distserve, patel2024splitwise, Hu-disaggregated}.

\begin{figure*}
    \centering
    \includegraphics[width=0.8\linewidth]{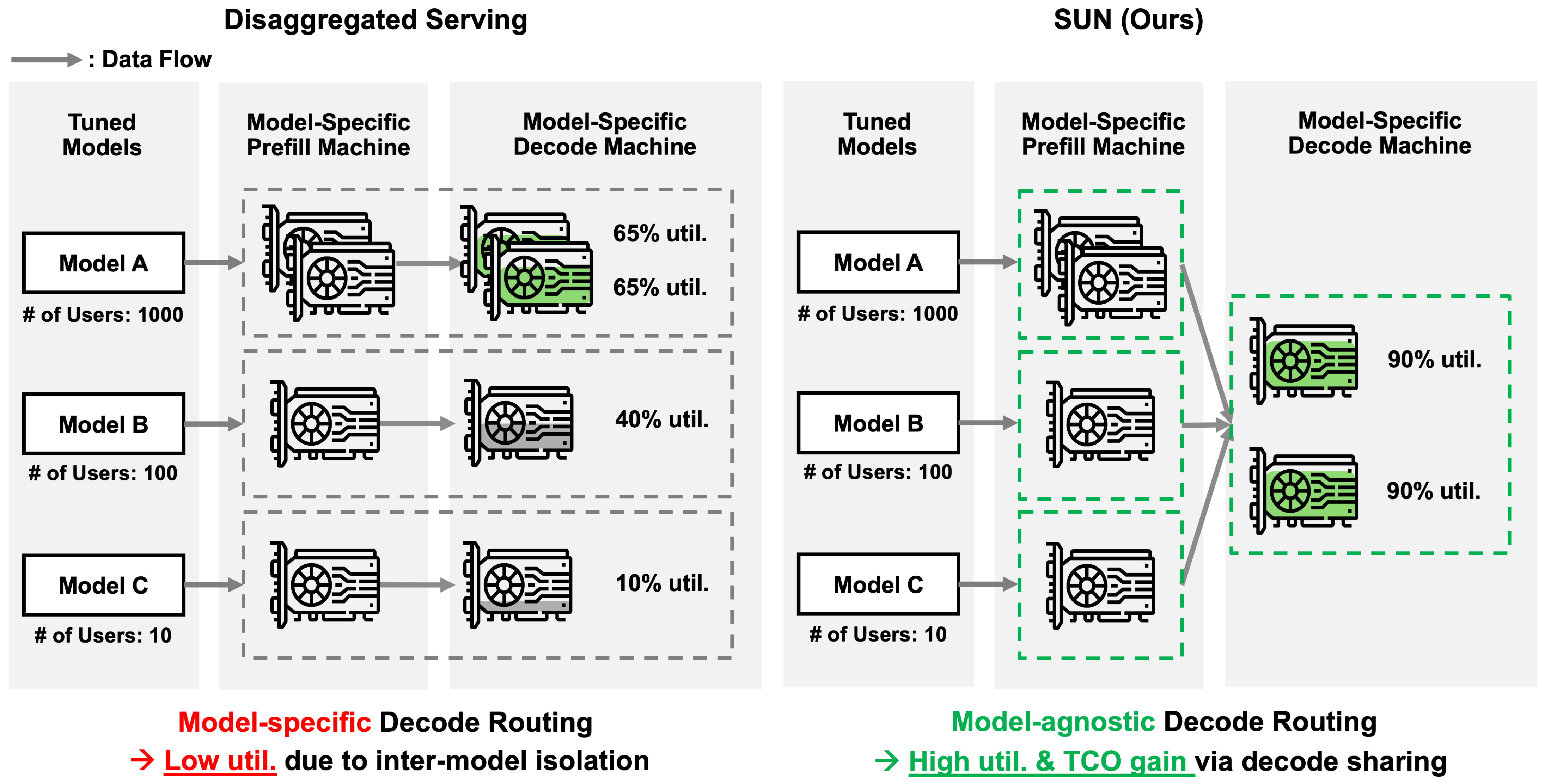}
    \caption{\textbf{Overview of SUN for disaggregated multi-LLM serving.}
 \textbf{(Left)}: In conventional disaggregated serving, decode workers are isolated per model, causing persistent GPU underutilization, especially for memory-bound decode execution under skewed multi-model workloads.
\textbf{(Right)}: SUN enables sharing a frozen decode module across different fine-tuned models by fine-tuning task-specific prefill modules.
This design improves GPU utilization, and reduces total cost of ownership (TCO).}
    \label{fig: SUN overview}
\end{figure*}

Disaggregated serving~\cite{Zhong-distserve,patel2024splitwise,Hu-disaggregated} mitigates this intra-model interference by separating prefill and decode onto different devices. For example, DistServe~\cite{Zhong-distserve} achieves up to 7.4$\times$ higher goodput while maintaining $>$90\% SLO attainment compared to a colocated baseline. However, while disaggregation addresses the interference within each model, this disaggregation does not resolve the inter-model isolation because decode workers remain partitioned on a per-model basis, as shown in the left panel of Fig.~\ref{fig: SUN overview}.

Inter-model isolation fragments capacity across models and prevents effective sharing of memory-bound decode resources. Although batching can improve decode utilization, cross-model batching is not possible when each model has its own dedicated decode workers. As a result, GPUs allocated to low-traffic models may remain underutilized while popular models queue decode requests, leading to persistent inefficiency. This issue is further exacerbated in real-world deployments~\cite{openrouter_fireworks_ondemand_2025,muxserve,Yao-deltazip}, where request distributions are highly skewed. 

To address these limitations, we propose Shared Use of Next-token Prediction (SUN), a novel algorithm that mitigates inter-model isolation in disaggregated multi-LLM serving. SUN decomposes a decoder-only LLM into a prefill module that produces KV caches and a decode module that generates tokens from them. For each downstream task, SUN fine-tunes only the prefill module while keeping the decode module frozen, enabling the same decode module to be shared across tasks. This design aligns training with decode sharing inference and enables model-agnostic decode routing to balance load across workers, improving GPU utilization. As a result, SUN reduces total cost of ownership (TCO) by consolidating memory-bound decode execution and better utilizing costly HBM-equipped GPUs, as depicted in Fig.~\ref{fig: SUN overview}.

We further propose Quantized SUN (QSUN) as a quantized variant of SUN. QSUN applies weight-only quantization to the shared decode module and performs prefill-only re-tuning to recover quantization-induced accuracy drop. By adapting only the prefill module after quantization, QSUN produces decoder-compatible KV caches while retaining low-bit decoding efficiency.

Experiments show that each specialized model with SUN matches full fine-tuning accuracy across diverse tasks and models, while maintaining system throughput with fewer decode GPUs in a vLLM-based disaggregated setup. SUN improves throughput per GPU by up to $2\times$ over a conventional disaggregated baseline while keeping TPOT within 5\%. QSUN further reduces TPOT by 45\% while recovering near full-precision SUN accuracy and outperforming AWQ in both accuracy and TTFT.

In summary, this work makes the following contributions:
\begin{itemize}
  \item \textbf{Robust cross-model decode sharing.} We propose SUN, the first algorithm that eliminates inter-model isolation in multi-LLM disaggregated serving by sharing a single frozen decode module across task-specific models via prefill-only tuning, preserving robust accuracy in various tasks and models.
  \item \textbf{Model-agnostic decode routing for high utilization.} We achieve efficient decode sharing by load-balancing decode requests across shared workers, maintaining overall throughput with 50\% fewer decode GPUs and incurring negligible latency overhead under skewed workloads in a vLLM-based disaggregated setup.
 \item \textbf{Accuracy-preserving quantized decoding.} We present QSUN which applies weight-only quantization to the shared decode module and performs prefill-only re-tuning, delivering a 45\% speedup while recovering near full-precision SUN accuracy and preserving the benefits of decode sharing.
\end{itemize}

\section{Background and Motivation}





\subsection{LLM Inference Optimization}
\label{sec:llm_inference_optimization}




\subsubsection{Continuous Batching}\label{sec: Continuous Batching}
Modern LLM serving systems aim to maximize GPU utilization under heterogeneous request lengths. While early systems relied on static batching, they suffered from poor utilization due to length variance across requests, as GPUs frequently became idle once shorter requests finished earlier than longer ones. Recent frameworks such as Orca~\cite{orca-yu}, vLLM~\cite{kwon-vllm}, and SGLang~\cite{sglnag-zheng} address this issue via continuous batching, which dynamically admits new requests and retires completed ones at each iteration, maintaining high GPU occupancy during autoregressive decoding.

\subsubsection{Disaggregated Serving}\label{sec:disaggregated_serving}
Continuous batching can still suffer from prefill--decode interference when both phases are co-located on the same device, increasing TPOT tail latency under high load. Disaggregated serving~\cite{Zhong-distserve, patel2024splitwise, Hu-disaggregated} mitigates this interference by executing prefill and decode on separate devices, enabling more stable decoding latency and better matching hardware to phase characteristics (compute-bound prefill vs.\ memory-bound decode). As a result, disaggregation has become a common design in modern LLM serving systems and is supported by frameworks such as vLLM~\cite{kwon-vllm} and SGLang~\cite{sglnag-zheng}. However, in multi-model deployments, disaggregation alone does not resolve inter-model isolation, since decode resources remain partitioned on a per-model basis.



\subsection{Motivation: Multi-LLM Serving Optimization}
\label{sec:motivation}

\subsubsection{Characteristics of Multi-LLM Serving}
\label{subsec:multi_model_serving}

Modern LLM deployments increasingly host multiple task- or domain-specialized models, rather than a single general-purpose model, driven by heterogeneous application demands and agentic workflows that orchestrate multiple specialists~\cite{muxserve, Yao-deltazip, Luo-skewedagnet1, yange-skewedagent2, wooldridge-skewedagent3}. A key characteristic of such deployments is strong traffic skew across models, resulting in a long-tail request distribution.

For example, OpenRouter statistics~\cite{openrouter_fireworks_ondemand_2025} show that the base model Llama-3.1-70B~\cite{llama3} serves on the order of $\sim$1B tokens per day, whereas fine-tuned variants receive substantially less traffic (e.g., $\sim$100M tokens per day for Nemotron-70B~\cite{nemotron} and Hermes-3-70B~\cite{ayed-heremes}). Similar skew also arises in hierarchical multi-agent systems~\cite{Luo-skewedagnet1,yange-skewedagent2,wooldridge-skewedagent3}, where a manager agent aggregates outputs from specialized agents and consequently bears a disproportionate fraction of requests and decoding load.


\subsubsection{Challenges in Multi-LLM Serving}
\label{subsec:multi_model_challenges}

Multi-LLM serving faces two key challenges: intra-model interference and inter-model isolation, as mentioned in Section~\ref{sec:introduction}. Intra-model interference stems from prefill-decode interference within a model pipeline, motivating disaggregated serving that separates the two phases~\cite{Zhong-distserve, patel2024splitwise, Hu-disaggregated}. However, the main remaining challenge is that GPUs cannot be shared across models. Since GPUs are allocated on a per-model basis, resources assigned to less popular models often stay underutilized, while popular models experience backlogs. This inefficiency is particularly pronounced during decoding: because decode is memory-bound, the execution tends to exhibit lower GPU utilization than the compute-bound prefill phase.

In single-model serving, mixed-request batching can mitigate underutilization by maintaining high GPU occupancy and amortizing per-step overhead. However, this strategy becomes less effective in multi-LLM serving, where decoding remains confined to model-specific worker pools and requests cannot be batch-merged across models. Consequently, the system must allocate separate decode GPUs per model, many of which operate at low utilization while others become backlogged especially under skewed demand~\cite{openrouter_fireworks_ondemand_2025,muxserve,Yao-deltazip}. From a TCO perspective, maintaining many underutilized GPUs is inefficient compared to cross-model sharing, which consolidates decoding onto a smaller pool of highly utilized GPUs and reduces overall GPU requirements.


\section{Methodology}
\label{sec:methodology}
In this section, we propose SUN, a novel disaggregated multi-LLM serving algorithm that mitigates inter-model isolation.
Section~\ref{subsec:formulation} formalizes decoder-only Transformers and introduces a prefill--decode decomposition aligned with LLM inference.
Section~\ref{subsec:sun} presents SUN, which preserves serving accuracy via prefill-only tuning and improves GPU utilization with model-agnostic decode routing.
Finally, Section~\ref{subsec:QSUN} introduces Quantized SUN (QSUN), which reduces TPOT while recovering quantization-induced accuracy via prefill re-tuning.

\subsection{Formulation}
\label{subsec:formulation}

\paragraph{Decoder-only Transformer}
A decoder-only Transformer defines an autoregressive distribution over output tokens conditioned on a prompt.
Given an input prompt $X=(x_1,\ldots,x_n)$ and previously generated tokens $y_{<t}=(y_1,\ldots,y_{t-1})$, a model with parameters $\theta$ defines the next-token distribution as
\begin{equation}
p_{\theta}(y_t \mid X, y_{<t}) \;=\; F_{\theta}(X, y_{<t}),
\label{eq:full_forward}
\end{equation}
where $F_\theta(\cdot)$ denotes the forward computation mapping the full context to a distribution over the next token. 

\paragraph{Prefill--Decode Decomposition}
To align the model structure with LLM inference scenario, we decompose a decoder-only Transformer in Eq.~\ref{eq:full_forward} into a prefill module and a decode module, such that the parameters of each module are only exercised in the corresponding inference phase
\begin{align}
\bigl(p_{\theta_p}(y_1 \mid X),\; \mathcal{C}_{X}\bigr) &= P_{\theta_p}(X), \label{eq:prefill}\\
\bigl(p_{\Theta}(y_{t} \mid X, y_{< t}),\; \mathcal{C}_{t}\bigr) &= D_{\theta_d}\!\left(y_{t-1},\; \mathcal{C}_{<t-1}\right), \label{eq:decode}
\end{align}
In Eqs.~\ref{eq:prefill}--\ref{eq:decode}, $\theta_p$ and $\theta_d$ denote the parameters used in the prefill and decode modules, respectively, and we define $\Theta \triangleq (\theta_p,\theta_d)$ as the full set of model parameters.

In Eq.~\ref{eq:prefill}, the prefill module $P_{\theta_p}$ processes the entire prompt $X=(x_1,\ldots,x_n)$ in parallel across prompt positions and returns (i) the next-token distribution for the first generation step and (ii) the prompt KV cache $\mathcal{C}_X$.
In Eq.~\ref{eq:decode}, the decode module $D_{\theta_d}$ is invoked once per step during autoregressive generation.
At step $t \ge 1$, the decode module consumes the most recent token $y_{t-1}$ together with the accumulated KV cache from previously processed positions $\mathcal{C}_{<t-1}$, and produces (i) the next-token distribution and (ii) the KV cache for the newly processed position $\mathcal{C}_{t}$.

This decomposition is purely conceptual: unless the modules are fine-tuned separately, both phases share the same base parameters, i.e., $\theta_p=\theta_d=\theta$.



Under this formulation, the KV cache up to step $t$ can be interpreted as
\begin{equation}
\mathcal{C}_{\le t}
\;=\;
\mathcal{C}_{X}\;\Vert\;\mathcal{C}_{y\leq t}.
\label{eq:cache_concat}
\end{equation}

In Eq.~\ref{eq:cache_concat}, $\mathcal{C}_X$ is produced by the prefill module $P_{\theta_p}(X)$, while $\mathcal{C}_{y\leq t}$ is incrementally generated by the decode module $D_{\theta_d}$ during autoregressive decoding.

\subsection{SUN: Shared Use of Next-token Prediction}
\label{subsec:sun}


\subsubsection{Concept of Decode Sharing}
\label{subsubsec:concept_of_decode_sharing}

Conventional fine-tuning produces task-specific models by updating a single parameter set that is used in both prefill and decode modules. Because each task-specific model has its own parameters (i.e., $\theta^{task_1} \neq \theta^{task_2}$), decode requests cannot be batch-merged across models, leading to substantial GPU underutilization in multi-LLM serving.

Here, we propose \textbf{Shared Use of Next-token Prediction (SUN)} to mitigate GPU underutilization due to inter-model isolation. SUN uses task-specific prefill modules (i.e., $\theta_p^{task_1} \neq \theta_p^{task_2}$) together with a single shared decode module, so all tasks decode with the same parameters $\theta_d^{base}$. While this design enables cross-model batching, two key questions remain.

\begin{itemize}
    \item \textbf{KV cache compatibility.} Can a shared base decode module with parameters $\theta_d^{\text{base}}$ correctly interpret KV caches $\mathcal{C}_X^{\text{task}}$ produced by task-specific prefill modules with parameters $\theta_p^{\text{task}}$ without sacrificing accuracy?
    \item \textbf{Decode routing in disaggregated serving.} How should decode requests be routed across shared decode workers, particularly in a disaggregated setup?
\end{itemize}

\begin{figure}
    \centering
    \includegraphics[width=0.65\linewidth]{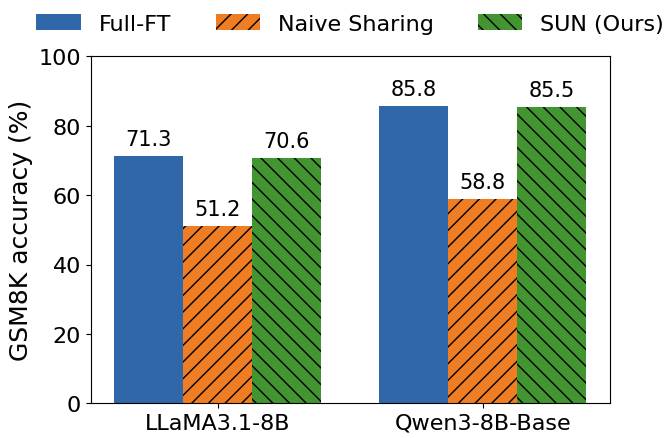}
    \caption{Impact of KV cache reuse strategies on GSM8K accuracy. Naive reuse of fine-tuned prefill caches with base model degrades accuracy significantly. In contrast, SUN matches full fine-tuning accuracy for LLaMA3.1-8B and Qwen3-8B-Base.}
    \label{fig:naive sharig}
\end{figure}

\subsubsection{Prefill-only tuning}\label{subsec:SUN}

\paragraph{Challenges in Sharing a Decoder Across Models.}
A straightforward approach to decoder sharing is to feed the KV cache generated by a task-specific fine-tuned prefill module into a frozen shared decode module. However, this naive sharing leads to severe accuracy degradation, because the base decoder is not trained to interpret KV caches produced under task-specific prefill parameters. As shown in Fig.~\ref{fig:naive sharig}, naive sharing incurs a substantial drop in GSM8K accuracy (e.g., from 71.3\% to 51.2\% on LLaMA3.1-8B, and from 85.8\% to 58.8\% on Qwen3-8B-Base), highlighting a significant train--inference mismatch.

\paragraph{Prefill-Only Tuning for Robust Decode Sharing.}
To enable the shared decode module to correctly interpret KV caches produced by task-specific prefill modules, we adopt prefill-only tuning. For each task $\tau$, we fine-tune only the task-specific prefill parameters $\theta_p^{\tau}$ (e.g., $\theta_p^{\text{task}_1}$, $\theta_p^{\text{task}_2}$, $\theta_p^{\text{task}_3}$) while keeping the shared decode parameters $\theta_d^{\text{base}}$ frozen. Given an input prompt $X$, the task-specific prefill module produces the prompt KV cache $\mathcal{C}_X^{\tau}$:
\begin{equation}
\bigl(p_{\theta_p^{\tau}}(y_1 \mid X),\, \mathcal{C}_{X}^{\tau}\bigr)=P_{\theta_p^{\tau}}(X).
\label{eq:prefill tau}
\end{equation}
The frozen decode module processes $y_{1:T}$ and computes the next-token distributions at all target positions, conditioning each position $t$ on $\mathcal{C}_X^{\tau}\Vert \mathcal{C}_{y_{<t}}$. We minimize the negative log-likelihood with respect to $\theta_p^{\tau}$:
\begin{equation}
\mathcal{L}(\theta_p^{\tau})
= - \sum_{t=1}^{T}
\log p_{\theta_d^{\text{base}}}\!\left(
y_t \mid y_{<t},\, \mathcal{C}_X^{\tau}\,\mathbin{\Vert}\,\mathcal{C}_{y_{<t}}
\right).
\label{eq:loss tau}
\end{equation}
Gradients from Eq.~\ref{eq:loss tau} backpropagate only to the prefill module $P_{\theta_p^{\tau}}$ through the prompt KV cache $\mathcal{C}_X^{\tau}$, while the shared decoder with parameters $\theta_d^{\text{base}}$ remains frozen. This training procedure encourages each task-specific prefill module to produce task-adaptive yet decoder-compatible KV caches, enabling robust decode sharing and achieving accuracy comparable to full fine-tuning (Fig.~\ref{fig:naive sharig}). Overall, prefill-only tuning aligns training with inference-time execution, eliminating train--inference discrepancies in decode sharing.

\subsubsection{Model-agnostic Routing Policy}
\label{subsubsec:inference_routing}

Efficient cross-model batching with a decode sharing requires an effective routing policy. Accordingly, we propose a model-agnostic routing policy, including model-specific prefill routing and model-agnostic decode routing.

\paragraph{Model-specific Prefill Routing.}
Incoming requests are deterministically routed to task-specific prefill modules. For example, math reasoning requests are directed to the math-tuned prefill module, while code generation requests are sent to the code-tuned prefill module. 

\paragraph{Model-agnostic Decode Routing.}
Unlike prefill routing, decode routing is decoupled from task identity. In particular, decode requests are dispatched to the shared decode worker pool in a model-agnostic manner, balancing load across workers regardless of per-model request skew.

This decoupling is the key enabler of cross-model batching: requests originating from different task-specialized prefill modules can be merged into a unified decode batch. Consequently, the system achieves high throughput with a small pool of highly utilized shared GPUs rather than many low-utilization model-specific GPUs, reducing the required GPU capacity and TCO. The shared pool sustains high utilization even when request arrival rates vary widely across models.

We implement this pipeline on top of the disaggregated serving setup in vLLM~\cite{kwon-vllm}, leveraging vLLM mechanisms for KV cache management and transfer between prefill and decode devices.


\begin{table*}[!t]
\centering
\caption{Accuracy comparison of full-fine tuning (Full-FT) and SUN with prefill-only tuning across diverse base models (LLaMA-3.1-8B and Qwen3-1.7B/8B/14B). For each domain, models are fine-tuned on MetaMathQA-40K (math), EvolInstruct-Code-80K (coding), and xLAM-function-calling-60K (tool calling), and evaluated on GSM8K/GSM, HumanEval/HumanEval+, and BFCL, respectively. Overall, SUN achieves accuracy comparable to full fine-tuning in most settings, while enabling shared decode execution across multiple models.}
\label{tab:main_results}
\small
\setlength{\tabcolsep}{6pt}

\begin{tabular}{l|c|cc cc cc}
\toprule
\multirow{2}{*}{\textbf{Model Configuration}} & \multirow{2}{*}{\textbf{Decode Execution}}
& \multicolumn{2}{c}{\textbf{Math}}
& \multicolumn{2}{c}{\textbf{Coding}}
& \multicolumn{2}{c}{\textbf{Tool calling}} \\
\cmidrule(lr){3-4}\cmidrule(lr){5-6}\cmidrule(lr){7-8}
& & \textbf{GSM8K} & \textbf{GSM+} & \textbf{HumanEval} & \textbf{HumanEval+} & \textbf{Simple Python} & \textbf{Multiple} \\
\midrule
LLaMA3.1-8B        & N/A & 25.9 & 18.0 & 36.6 & 29.9 & 70.5 & 45.5 \\
\cdashline{1-8}\noalign{\vskip 0.35ex}
+Full-FT            & Independent & \textbf{71.3} & \textbf{49.8} & 48.2 & \textbf{45.7} & \textbf{90.0} & 88.0 \\
+SUN (Ours) & \textbf{Shared} & 70.6 & 49.3 & \textbf{49.4} & 41.5 & 89.8 & \textbf{90.5} \\
\midrule

Qwen3-1.7B-Base     & N/A &  9.9 &  9.8 & 48.2 & 39.6 & 82.5 & 60.0 \\
\cdashline{1-8}\noalign{\vskip 0.35ex}
+Full-FT             & Independent & \textbf{75.0} & \textbf{54.8} & \textbf{67.1} & \textbf{59.8} & \textbf{89.8} & \textbf{90.0} \\
+SUN (Ours)  & \textbf{Shared} & 74.1 & 53.9 & \textbf{67.1} & 58.5 & 88.8 & 89.0 \\
\midrule

Qwen3-8B-Base        & N/A & 11.8 & 12.5 & 68.3 & 61.6 & 81.5 & 80.0 \\
\cdashline{1-8}\noalign{\vskip 0.35ex}
+Full-FT              & Independent & \textbf{85.8} & \textbf{65.7} & 83.5 & 74.3 & \textbf{93.3} & \textbf{92.0} \\
+SUN (Ours)   & \textbf{Shared} & 85.5 & 64.5 & \textbf{84.1} & \textbf{76.2} & \textbf{93.3} & 91.5 \\
\midrule

Qwen3-14B-Base       & N/A & 25.9 & 24.5 & 68.9 & 62.3 & 88.0 & 77.0 \\
\cdashline{1-8}\noalign{\vskip 0.35ex}
+Full-FT              & Independent & \textbf{88.1} & \textbf{67.5} & 85.3 & 78.0 & \textbf{93.8} & \textbf{92.5} \\
+SUN (Ours)   & \textbf{Shared} & 85.4 & 64.6 & \textbf{88.4} & \textbf{79.9} & 93.5 & \textbf{92.5} \\

\bottomrule
\end{tabular}
\end{table*}

\subsection{QSUN: Decode Module Quantization in SUN}\label{subsec:QSUN}

In a disaggregated serving setup, prefill and decode are executed on separate devices, enabling phase-specific optimizations. Since prefill is compute-bound, weight-only quantization typically provides limited latency benefits and can even increase latency due to dequantization overhead. In contrast, decode is memory-bound, making it the primary phase where quantization can yield substantial efficiency gains. Motivated by this observation, we selectively apply weight-only quantization only to the decode module:
\begin{align}
\bigl(p_{\theta_p^{\tau}}(y_1 \mid X),\; \mathcal{C}_{X}\bigr) &= P_{\theta_p^{\tau}}(X), \label{eq:qprefill}\\
\bigl(p_{\Theta}(y_{t} \mid X, y_{< t}),\; \mathcal{C}_{t}\bigr) &= D_{q(\theta_d)}\!\left(y_{t-1},\; \mathcal{C}_{<t-1}\right), \label{eq:qdecode}
\end{align}
where $q(\theta_d)$ denotes weight-only quantization applied to the decode parameters $\theta_d$.
This design maximizes the latency and throughput benefits of quantization by targeting the memory-bound decoding bottleneck while keeping the prefill phase in high precision. However, naively quantizing the decode module can still cause non-trivial accuracy degradation due to a representation mismatch between the high-precision prefill module and the low-precision decode module.


To mitigate quantization error, we propose Quantized SUN (QSUN), which performs prefill-only re-tuning after quantizing the shared decode module. Specifically, the quantized decode module $D_{q(\theta_d)}$ in Eq.~\ref{eq:qdecode} is kept frozen, and only the task-specific prefill parameters $\theta_p^{\tau}$ in Eq.~\ref{eq:qprefill} are updated. Under this procedure, the prefill module learns to produce task-specific KV representations that remain compatible with the shared low-bit decoder. Please note that we do not consider re-tuning the decode module to recover quantization error, as updating $\theta_d$ would break decode sharing and invalidate cross-model request redistribution. Through re-tuning, we recover the accuracy degradation caused by quantizing the decode module in SUN.

From a systems perspective, QSUN combines high-precision task-specific prefill modules with a low-precision shared decode module, matching the compute characteristics of the two phases. Specifically, keeping prefill in full precision avoids quantization-induced computation overhead and preserves time to first token (TTFT), while weight-only quantization of the shared decode module reduces weight traffic during decoding and improves TPOT.

\section{Experimental Results}

\subsection{Experimental setup}

\label{subsec:exp_setup}

\paragraph{Training Setup}
We compared full fine-tuning (Full-FT) and SUN on LLaMA3.1-8B \cite{llama3} and Qwen3-Base models (1.7B/8B/14B) \cite{yang2025qwen3}.
For domain adaptation, the models are trained on MetaMathQA-40K \cite{yu2023metamath}, EvolInstruct-Code-80K \cite{evolinstruct}, and xLAM-function-calling-60K \cite{XLAM}, targeting math, code generation, and tool use, respectively. We then evaluate each adapted model on its corresponding benchmark: GSM8K/GSM+ \cite{cobbe2021gsm8k,li-etal-2024-gsmplus} (math), HumanEval/HumanEval+ \cite{chen2021humaneval,liu2023evalplus} (coding), and BFCL \cite{patil2025bfcl} (tool calling), using the Simple Python and Multiple subsets. The details are provided in Appendix \ref{app:training_setup}.

\paragraph{Inference Setup}
We consider a multi-LLM serving setup on a single DGX A100 node equipped with 8 GPUs, hosting four LLaMA3.1-8B-based models. The baseline adpots a disaggregated serving configuration that assigns 1 prefill GPU and 1 decode GPU per model instance (4$\times$1P/1D). In contrast, SUN allocates 1 prefill GPU per model but pools the $N$ decode GPUs across models (4P/$N$D).

We evaluate whether SUN can maintain baseline system throughput with fewer decode GPUs. Under a fixed offered request rate (RPS), we vary the output sequence length (OSL) and sweep the number of decode GPUs from 1 to 4. To study skewed workloads, we distribute per-model RPS according to a Zipf distribution with skew parameter $\alpha$ (larger $\alpha$ indicates more imbalanced traffic across models), following prior work~\cite{Yao-deltazip}. Detailed experimental settings are provided in Appendix~\ref{app:inference_setup}.

\begin{figure*}
    \centering
    \includegraphics[width=0.90\linewidth]{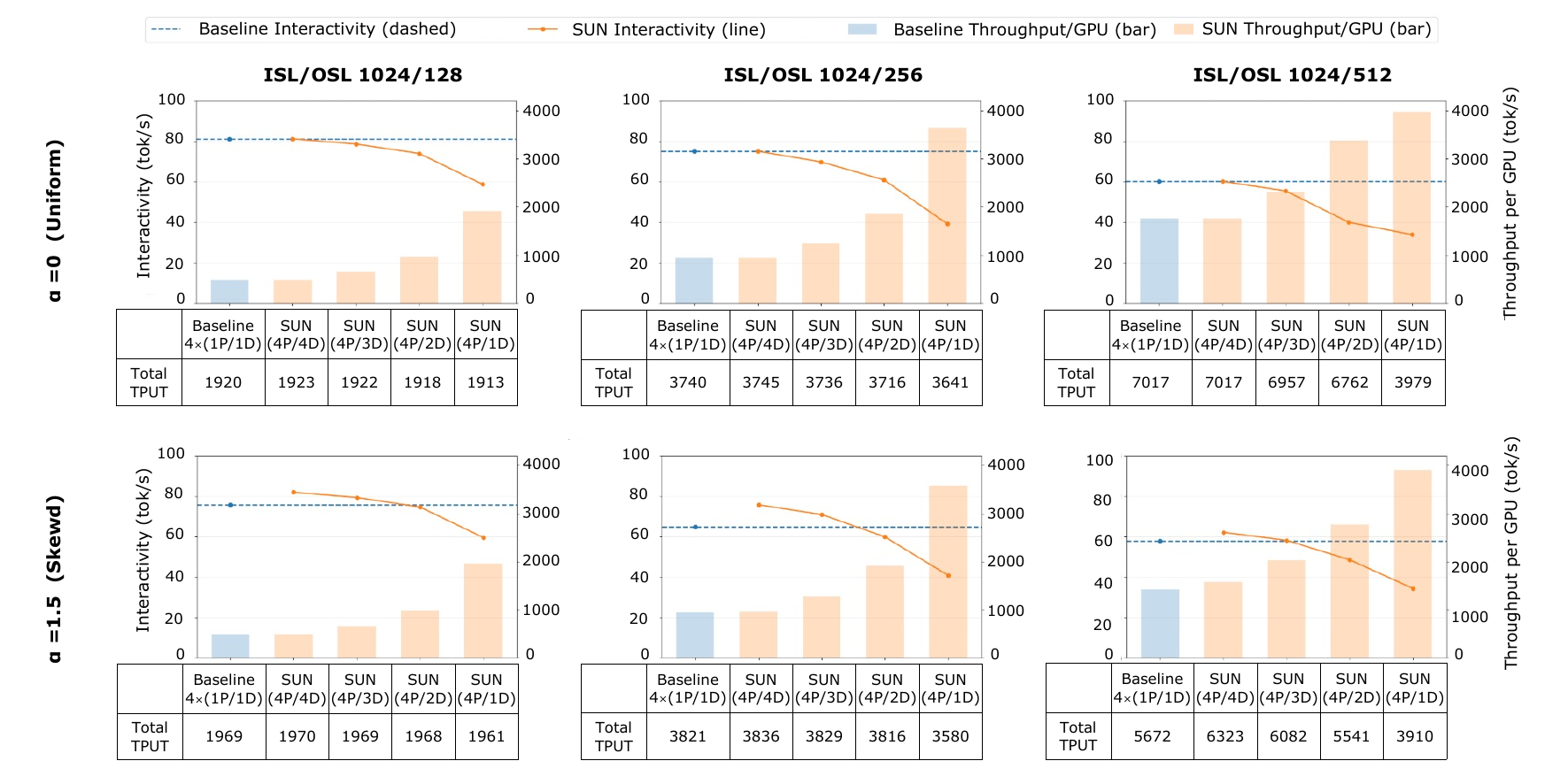} 
    \caption{Throughput--interactivity trade-off under decode-GPU consolidation.
Comparison of a per-model partitioned baseline (4$\times$(1P/1D)) with SUN, which uses task-specific prefill workers (4P) and shares a variable number of decode workers (4D/3D/2D/1D), under uniform ($\alpha{=}0$) and skewed ($\alpha{=}1.5$) request distributions and varying output sequence lengths (OSL).
Bars indicate throughput per GPU and lines indicate interactivity ($=1/\mathrm{TPOT}$); the table reports total system throughput (TPUT).
Near-constant total throughput is maintained when consolidating decode GPUs down to 2D, revealing a controllable throughput-latency trade-off under skewed workloads.}
    \label{fig: inf-main-results}
\end{figure*}

\begin{figure}[!h]
  \centering
  \includegraphics[width=0.70\columnwidth]{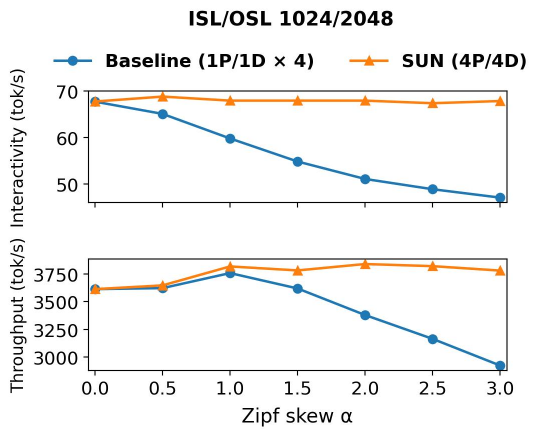}
  \caption{Effect of workload skew on interactivity and throughput.
Baseline (blue) partitions resources per model, where each of the four models uses one dedicated prefill GPU and one dedicated decode GPU (1P/1D $\times$ 4).
SUN (orange) assigns four task-specific prefill GPUs and shares four decode GPUs across models (4P/4D).
As the Zipf skew $\alpha$ increases, SUN maintains stable throughput and interactivity, demonstrating robustness to request imbalance.}
  \label{fig:zipf_throughput}
\end{figure}

\subsection{Accuracy of Prefill-Only Tuning}
As discussed in Section~\ref{subsubsec:concept_of_decode_sharing}, one of our key questions is whether a shared base decode module can reliably interpret the KV caches produced by task-specific prefill modules. 
Table~\ref{tab:main_results} answers this question affirmatively, showing that SUN enables shared decode execution while preserving Full-FT-level accuracy across models and tasks.

On LLaMA3.1-8B, SUN matches Full-FT on most benchmarks and even improves performance on HumanEval (49.4 vs.\ 48.2) and BFCL-Multiple (90.5 vs.\ 88.0). 
For Qwen3-1.7B-Base and Qwen3-8B-Base, SUN similarly maintains accuracy close to Full-FT across math, coding, and tool-calling tasks, with notable gains on Qwen3-8B-Base for coding (HumanEval: 84.1 vs.\ 83.5; HumanEval+: 76.2 vs.\ 74.3). 
While SUN shows a moderate drop on math for Qwen3-14B-Base (GSM8K: 85.4 vs.\ 88.1; GSM+: 64.6 vs.\ 67.5), it improves coding accuracy (HumanEval: 88.4 vs.\ 85.3; HumanEval+: 79.9 vs.\ 78.0). 


Overall, these results suggest that prefill-only tuning produces task-adaptive KV caches that remain compatible with a shared frozen decoder, enabling robust decode sharing across diverse tasks and model scales. 
We next quantify the system-level efficiency gains of SUN in Section~\ref{subsec:latency_improvement}.


\subsection{Performance under Decoder Sharing}
\label{subsec:latency_improvement}


\textbf{Decode-GPU Consolidation and TCO Benefit}

A key advantage of SUN is that it aggregates decode requests from heterogeneous models and serves them on a shared pool of decode GPUs, improving utilization and enabling decode-GPU consolidation. As shown in Fig.~\ref{fig: inf-main-results}, throughput per decode GPU increases as the decode pool shrinks. This consolidation, however, introduces an inherent throughput--latency trade-off: reducing decode GPUs can create decode-side bottlenecks and lower interactivity, especially at longer OSL. 

For instance, under OSL$=256$ and uniform workload (i.e. $\alpha=0$), throughput per decode GPU increases from 935 tok/s (baseline) to 1245, 1858, and 3641 tok/s as we reduce the number of decode workers. This gain comes at the cost of interactivity: compared with the baseline, interactivity degrades by 7\%, 19\%, and 48\%, respectively. Notably, SUN can reduce the decode pool to 3 workers while maintaining total system throughput, achieving a 33\% increase in throughput per GPU with only a 7\% TPOT degradation relative to the baseline setup.

The advantage of SUN becomes more pronounced under skewed workloads, where decode demand is highly imbalanced across models (e.g., a small subset of models dominates decode traffic while others are only sporadically invoked). In such settings, the baseline can suffer from decode-worker underutilization since model-specific decode pipelines cannot efficiently absorb load imbalance. By contrast, SUN enables effective load balancing across decode workers by adopting a single decode module shared across all task-specific models, allowing decode capacity to be flexibly allocated to whichever requests are active.

Under skewed workloads ($\alpha=1.5$), SUN maintains comparable total system throughput and achieves up to $2.0\times$ higher throughput per decode GPU, while keeping interactivity degradation within 10\% even when reducing the decode pool to 2 workers for OSL$=128$ and 256. For the decode-heavy setting (OSL$=512$), SUN further improves total system throughput by 11\% over the baseline with 3 decode workers (vs.\ 4), while limiting TPOT degradation to within 8\%.



\begin{table*}[!t]
\centering
\caption{System performance and accuracy comparison of Full-FT, AWQ, and QSUN on LLaMA3.1-8B and Qwen3-8B-Base. We report time-to-first-token (TTFT, ms), time-per-output-token (TPOT, ms), throughput (TPUT, tok/s), and accuracy on math/coding benchmarks. Across both models, QSUN consistently improves accuracy over AWQ while preserving the 4-bit decode efficiency (TPOT/TPUT) relative to the 16-bit Full-FT baseline. Moreover, by avoiding quantized prefill overhead, QSUN achieves lower TTFT than AWQ, while retaining SUN's shared low-bit decoder execution across tasks.}
\label{tab: qsun-results}
\small
\begin{tabular}{c|c|c|c|ccc|cc|cc}
\toprule
\multirow{2}{*}{\textbf{Model}} &
\multirow{2}{*}{\textbf{Method}} &
\multirow{2}{*}{\textbf{\# Bits}} &
\multirow{2}{*}{\textbf{\makecell{Decode\\Execution}}} &
\multicolumn{3}{c|}{\textbf{System Performance}} &
\multicolumn{2}{c|}{\textbf{Math}} &
\multicolumn{2}{c}{\textbf{Coding}} \\
\cmidrule(lr){5-7}\cmidrule(lr){8-9}\cmidrule(lr){10-11}
& & & &
\textbf{TTFT} & \textbf{TPOT} & \textbf{TPUT} &
\textbf{GSM8K} & \textbf{GSM+} &
\textbf{HEval} & \textbf{HEval+} \\
\midrule
\multirow{3}{*}{\textbf{LLaMA3.1-8B}}
& Full-FT & 16/16 & Independent & 99.1 & 13.8 & 73.3  & 71.3 & 49.8 & 48.2 & 45.7 \\
& +AWQ     & 4/4   & Independent & 118.7 & \textbf{7.6} & \textbf{130.8} & 70.2 & 47.6 & 45.7 & 39.6 \\  \cdashline{2-11}\noalign{\vskip 0.35ex}
& QSUN    & 16/4  & \textbf{Shared} & \textbf{98.9} & \textbf{7.6} & 130.2 & \textbf{70.3} & \textbf{48.1} & \textbf{47.6} & \textbf{42.7} \\
\midrule
\multirow{3}{*}{\textbf{Qwen3-8B-Base}}
& Full-FT & 16/16 & Independent & 96.4 & 12.2 & 81.3 & 85.8 & 65.7 & 83.5 & 74.3 \\ 
& +AWQ     & 4/4   & Independent & 115.8 & 6.6  & 148.9 & 83.6 & 64.0 & 79.3 & 72.6 \\ \cdashline{2-11}\noalign{\vskip 0.35ex}
& QSUN    & 16/4  & \textbf{Shared} & \textbf{97.1} & \textbf{6.6}  & \textbf{147.9} & \textbf{85.1} & \textbf{64.9} & \textbf{82.3} & \textbf{75.0} \\
\bottomrule
\end{tabular}%
\end{table*}

\textbf{Robustness to Skewed Workloads}
Figure~\ref{fig:zipf_throughput} illustrates how workload skew affects serving performance in terms of interactivity and throughput. Increasing skew leads to a substantial degradation in interactivity, since requests concentrate on a small subset of models and create persistent bottlenecks on their dedicated decode resources. For the same reason, total system throughput decreases monotonically as $\alpha$ increases, dropping by up to 19.0\% relative to the uniform workload ($\alpha=0$) under extreme skew ($\alpha=3.0$).

In contrast, SUN maintains stable interactivity and throughput even under highly skewed workloads. Even under a severely skewed workload ($\alpha=3.0$), SUN achieves interactivity and throughput comparable to the uniform-workload setting, improving interactivity by 44.1\% and total system throughput by 29.2\% relative to the $\alpha=3.0$ baseline. By sharing decode resources across models, SUN can redistribute decode requests to available workers, preventing bottlenecks caused by per-model resource partitioning. These results highlight that decode sharing is particularly effective in realistic multi-LLM serving scenarios where request distributions are highly imbalanced.

\subsection{Results of Quantized SUN}

\paragraph{Accuracy of QSUN} In Table \ref{tab: qsun-results}, we observe that QSUN achieves accuracy close to the 16-bit Full-FT baselines, with only modest degradation, despite sharing a single low-precision (4-bit) decode module across multiple models. For LLaMA3.1-8B, QSUN remains within 1\% of Full-FT on GSM8K and HumanEval. Although performance drops slightly for GSM+ and HumanEval+, QSUN still surpasses AWQ under the same 4-bit decoding setting by 0.5\% and 3.1\%, respectively. The same pattern holds for Qwen3-8B-Base, where QSUN consistently outperforms AWQ across all tasks. These results suggest that prefill-only re-tuning mitigates the discrepancy between the prefill and the low-precision decode module. Further analysis on the effectiveness of prefill-only re-tuning is provided in Appendix~\ref{app:recover}.






\paragraph{Latency and Throughput.}
Table \ref{tab: qsun-results} further shows that QSUN outperforms AWQ in system performance across both model families. Specifically, on LLaMA3.1-8B and Qwen3-8B-Base, QSUN achieves TPOT and output-token throughput comparable to AWQ while substantially improving over the 16-bit Full-FT baseline, reducing TPOT by up to 45\% and increasing throughput by up to $1.7\times$.
 Moreover, because QSUN keeps the compute-bound prefill stage in full precision and avoids unnecessary quantization/dequantization overhead, it reduces TTFT by 17\% relative to AWQ, bringing TTFT back to a level comparable to the Full-FT baseline. This trend persists at higher concurrency and input sequence length, as shown in Appendix~\ref{app: qsun performance}.

\section{Related Works}

\subsection{Disaggregated LLM Inference.} Disaggregated prefill and decode phases has emerged as a dominant paradigm in production LLM serving. Seminal works, such as Splitwise~\cite{patel2024splitwise}, DistServe~\cite{Zhong-distserve}, and Mooncake~\cite{qin2024mooncake}, demonstrate that phase-level disaggregation effectively resolves resource contention and maximizes throughput. Our work builds upon this established foundation but fundamentally extends its scope: we propose decoder sharing across models to address resource fragmentation in multi-LLM settings—a critical challenge not addressed by prior work.

\subsection{Multi-LLM Serving Architectures}
Serving multiple specialized models concurrently introduces challenges in GPU utilization. Early systems such as AlpaServe~\cite{Alphaserve} rely on model multiplexing by swapping weights between host memory and GPU, but the weight-loading overhead limits interactivity. More recent Multi-LoRA serving approaches (e.g., Punica~\cite{punica} and S-LoRA~\cite{slora}) reduce memory cost by sharing a frozen base model across adapters, but require specialized runtime support and incur decoding overhead due to adapter-specific fragmentation. In contrast, SUN shares the exact a single decode module across models without runtime adapter switching, enabling efficient multi-LLM serving with standard decoding kernels.


\section{Conclusion}
In this work, we address a key challenge in disaggregated multi-LLM serving: inter-model isolation arising from model-specific partitioning of memory-bound decoding resources. We propose SUN, which removes this isolation by sharing a common decode module across heterogeneous models while fine-tuning task-specific prefill modules. We show that prefill-only tuning aligns each prefill module to produce KV caches that are directly interpretable by a shared decoder, enabling shared decode execution without sacrificing the accuracy of fully fine-tuned models. By pooling decode workers and flexibly redistributing capacity under skewed multi-LLM workloads, SUN improves utilization and can reduce TCO in multi-LLM disaggreagated serving.

\section*{Impact Statement}

This work introduces SUN to improve the efficiency of multi-LLM serving. By reducing GPU underutilization during decoding, our approach can lower the cost and energy required to serve multiple specialized models, making LLM-based applications more accessible.


\nocite{langley00}

\bibliography{example_paper}
\bibliographystyle{icml2026}

\newpage
\appendix
\onecolumn
\section*{Appendices}
\section{Experimental Setup}\label{app:experimental setup}
\subsection{Training Setup}\label{app:training_setup}

\paragraph{Prefill-Only Tuning Configurations.}
For prefill-only tuning of SUN, we fine-tune the task-specific prefill module while keeping the shared decoder frozen.
We use the AdamW optimizer with a linear learning rate scheduler and a warmup ratio of 0.1. 
We perform a learning rate search over \{2e{-}6, 5e{-}6, 1e{-}5, 2e{-}5, 5e{-}5\}, and report the best results.
The global batch size is set to 128, and models are trained for a single epoch, which we find sufficient for convergence.

\paragraph{Quantization Configurations.}
For Quantized SUN, we apply weight-only post-training quantization to the shared decoder using AWQ~\cite{awq}.
We implement AWQ with the LLM Compressor library~\cite{llmcompressor2024} and follow its standard configuration without modifying the quantization algorithm.
All linear layers in the decoder are quantized to 4-bit precision using per-channel, symmetric quantization with a group size of 128, except for the \texttt{lm\_head} layer, which remains in full precision.
Calibration is performed using 256 samples randomly selected from the HuggingFaceH4/ultrachat\_200k dataset~\cite{ding2023enhancing}, with a maximum sequence length of 512, following the default practice of LLM Compressor.

\paragraph{Quantized SUN Re-tuning Procedure and Configurations.}
After applying decoder quantization, we perform a encoder re-tuning step following the same prefill-only tuning setup as SUN.
Specifically, we re-tune the prefill encoder for one additional epoch using the same training dataset and objective, while keeping the quantized decoder frozen.
We use the same optimizer and scheduler configuration as in prefill-only tuning, and do not introduce additional regularization.
This re-tuning step is sufficient to recover most of the accuracy degradation introduced by decoder quantization.







\subsection{Inference Setup}
\label{app:inference_setup}

\paragraph{Testbed and serving stack.}
We run all inference experiments on a single DGX A100 node with 8 NVIDIA A100 GPUs.
We serve $N_{\text{model}}{=}4$ model instances (referred to as \emph{agents}) using a disaggregated prefill--decode serving architecture built on vLLM~\cite{kwon-vllm}.
Each agent corresponds to one deployed model instance (a task-specialized fine-tuned variant of the same backbone with identical architecture and size).
Each request is split into a prefill phase (processing the input prompt and producing KV cache) and a decode phase (streaming token generation from the KV cache).

\paragraph{Serving configurations.}
We compare two configurations.
\emph{Baseline (model-isolated 4$\times$1P/1D):} each agent is assigned one dedicated prefill GPU and one dedicated decode GPU (4 prefill + 4 decode GPUs).
\emph{SUN (decoder pooling 4P/$K$D):} each agent keeps a dedicated prefill GPU, while decode GPUs are pooled across agents with pool size
$K \in \{1,2,3,4\}$.
When $K<4$, SUN consolidates decode capacity while keeping the prefill footprint unchanged.

\paragraph{Routing policy.}
Prefill routing is agent-specific: each request is deterministically routed to its corresponding agent prefill worker.
Decode routing is configuration-dependent: the baseline routes decode to a fixed per-agent decode worker (model-isolated),
whereas SUN dispatches decode requests using a model-agnostic load distribution rule across the shared decode pool of size $K$,
independent of agent identity. This isolates the effect of pooling from model-specific scheduling heuristics.

\paragraph{Workloads and offered load.}
We use streaming generation with fixed input sequence length (ISL) and target output sequence length (OSL).
To ensure comparable decode work across configurations, we enforce the generated output length to stay close to the target OSL and verify this empirically; for example, with OSL$=512$ we observe a mean realized output length of 511.8 tokens with std.\ 0.5 (min 509, max 513).
We control the offered request rate per agent at the client and define the offered total request rate as the sum across agents.

\paragraph{Skewed workloads (Zipf).}
To emulate traffic skew in multi-LLM serving, we distribute the offered total request rate $R_{\text{total}}$ across the four agents
according to a Zipf distribution~\cite{Yao-deltazip}:
\[
P(i) = \frac{i^{-\alpha}}{\sum_{j=1}^{N_{\text{model}}} j^{-\alpha}}, \qquad
R_i = R_{\text{total}} \cdot P(i),
\]
where $i$ is the popularity rank and $\alpha$ controls skew ($\alpha{=}0$ is uniform).

\paragraph{Caching configuration.}
We disable prefix caching for all experiments to avoid cache-hit artifacts in TTFT/TPOT and to ensure that results reflect scheduling and pooling effects rather than prompt reuse.

\paragraph{Metrics.}
We report output-token throughput (tok/s) and token-level latency metrics from the streaming client: TTFT, ITL, and TPOT.
We define \emph{interactivity} as the inverse of TPOT, $I = 1/\mathrm{TPOT}$ (token/ms).
For readability, we report interactivity in tok/s as $I_{\text{tok/s}} = 1000/\mathrm{TPOT}_{\text{ms}}$ and use this definition consistently across figures.

\paragraph{Measurement protocol.}
Each experiment runs continuously with an initial grace period, followed by a fixed measurement window.
Unless stated otherwise, we use a 30\,s grace period and a 60\,s measurement window.
For long-output settings (e.g., ISL=1024, OSL=2048), we extend the window (e.g., 60\,s grace + 180\,s measurement) to reduce variance by collecting more completed requests.
All metrics are computed over requests completed within the measurement window.

\begin{figure}[t]
\centering
\begin{subfigure}[t]{0.48\linewidth}
\centering
\includegraphics[width=\linewidth]{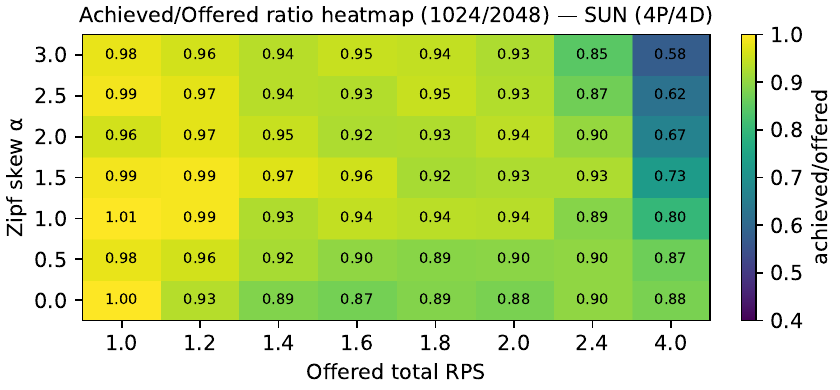}
\caption{SUN (shared decode pool)}
\label{fig:ach_off_heatmap_sun}
\end{subfigure}\hfill
\begin{subfigure}[t]{0.48\linewidth}
\centering
\includegraphics[width=\linewidth]{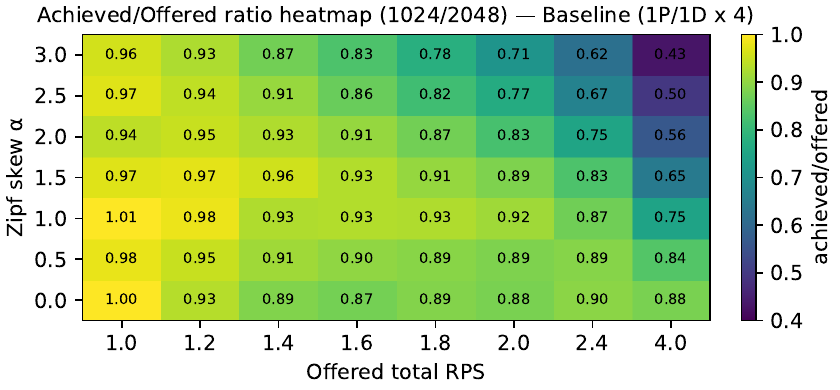}
\caption{Baseline (model-isolated)}
\label{fig:ach_off_heatmap_baseline}
\end{subfigure}

\caption{Achieved/offered request-rate ratio (ISL=1024, OSL=2048) across Zipf skew $\alpha$ and offered total RPS.
Ratios close to 1 indicate that the system sustains the injected load; lower ratios indicate overload/backlog.
At the operating point used in the main skew sweep (offered total RPS$=2$), the ratio is generally below 1 and degrades with increasing $\alpha$,
indicating incipient backlog under skew; SUN remains consistently closer to 1 than the baseline.}
\label{fig:ach_off_heatmap}
\end{figure}

\paragraph{Achieved vs.\ offered load sanity check.}
To contextualize the Zipf-$\alpha$ sweep at offered total RPS$=2$, we report the achieved/offered request-rate ratio
over a grid of offered total RPS and Zipf skew $\alpha$ (Appendix Fig.~\ref{fig:ach_off_heatmap}).
We define the ratio as the measured achieved request rate divided by the offered total request rate;
values below 1 indicate that the system does not fully sustain the injected open-loop load due to queueing/backlog
(and, under severe overload, possible drops).
At the operating point used in the main skew sweep (ISL$=1024$, OSL$=2048$, offered total RPS$=2$), the ratio can fall below 1 under skew,
indicating a near-saturation regime with incipient backlog.
As $\alpha$ increases, the model-isolated baseline drops sharply
and falls further below 1, whereas SUN remains consistently closer to 1, implying that it sustains a larger fraction of the offered request rate under heavy skew. 
Since the realized output length stays close to the target OSL in our runs, throughput differences largely reflect differences in achieved request rate rather than output-length drift.
Overall, Appendix Fig.~\ref{fig:ach_off_heatmap} supports interpreting the skew-sweep trends
as robustness to skew at the same offered load, rather than artifacts of extreme overload.

\clearpage

\section{Effectiveness of Prefill-Only Re-tuning in QSUN}
\label{app:recover}

Table~\ref{tab:QSUN-perf-acc-app} presents an ablation study that isolates the impact of the proposed re-tuning step in Quantized SUN (QSUN). 
Recall that \textbf{SUN} fine-tunes only the task-specific prefill module while keeping a shared base decoder frozen, enabling decoder sharing across tasks. 
We compare the following variants:
(i) \textbf{Full-FT} (16/16), a standard fully fine-tuned model in full precision;
(ii) \textbf{Full-FT+AWQ} (4/4), where weight-only 4-bit AWQ quantization is applied to both prefill and decode modules;
(iii) \textbf{SUN} (16/16), which applies prefill-only tuning with a frozen decoder;
(iv) \textbf{SUN+AWQ-d} (16/4), which applies decoder-only AWQ quantization after SUN training; and
(v) \textbf{QSUN} (16/4), which further performs prefill-only re-tuning on top of SUN+AWQ-d.
We report accuracy on math and coding benchmarks (GSM8K, GSM+, HumanEval, and HumanEval+) for LLaMA3.1-8B and Qwen3-8B-Base.

The results highlight that decoder-only quantization without re-tuning can severely degrade accuracy. In particular, SUN+AWQ-d leads to a dramatic collapse on math for LLaMA3.1-8B (e.g., 11.4 on GSM8K and 7.0 on GSM+), indicating that naively quantizing the shared decoder breaks the compatibility between the task-specific prefill module and the shared decode module. This degradation is expected: SUN trains the prefill module to produce KV caches that are well-aligned with the full-precision frozen decoder, and post-hoc quantization introduces a representation mismatch that disrupts this learned adaptation.

QSUN resolves this issue through prefill-only re-tuning, which updates only the task-specific prefill module while keeping the quantized decoder frozen. This step encourages low-precision decoding-aware KV cache generation, allowing the prefill module to produce KV caches that the quantized shared decoder can reliably consume. As a result, QSUN substantially recovers the quantization-induced accuracy loss, reaching performance close to the pre-quantization SUN level and comparable to Full-FT. For example, on LLaMA3.1-8B, QSUN restores GSM8K accuracy from 11.4 to 70.3 and GSM+ from 7.0 to 48.1. Similar recovery is observed for Qwen3-8B-Base, where QSUN improves GSM8K from 79.8 to 85.1 and GSM+ from 60.4 to 64.9.

Overall, this ablation confirms that re-tuning is essential for preserving accuracy when applying decoder-only quantization under SUN, enabling QSUN to combine the system benefits of low-bit shared decoding with near full-precision task performance.

\begin{table*}[!t]
\centering
\caption{Accuracy comparison of Full-FT, Full-FT+AWQ, SUN, SUN+AWQ-d (decoder-only AWQ) and QSUN on LLaMA3.1-8B and Qwen3-8B-Base. We report accuracy on math/coding benchmarks. The \# Bits column denotes the precision (in bits) used in the prefill/decode phases (prefill/decode). \textbf{Bold} indicates QSUN results.}
\label{tab:QSUN-perf-acc-app}
\small
\begin{tabular}{c|c|c|c|c|c|c}
\toprule
\multirow{2}{*}{\textbf{Model}} &
\multirow{2}{*}{\textbf{Method}} &
\multirow{2}{*}{\textbf{\# Bits}} &
\multicolumn{2}{c|}{\textbf{Math}} &
\multicolumn{2}{c}{\textbf{Coding}} \\
\cmidrule(lr){4-5}\cmidrule(lr){6-7}
& & &
\textbf{GSM8K} & \textbf{GSM+} &
\textbf{HEval} & \textbf{HEval+} \\
\midrule
\multirow{3}{*}{\textbf{LLaMA3.1-8B}}
& Full-FT & 16/16 & 71.3 & 49.8 & 48.2 & 45.7 \\
& Full-FT+AWQ & 4/4   & 70.2 & 47.6 & 45.7 & 39.6 \\  \cdashline{2-7}\noalign{\vskip 0.35ex}
& SUN & 16/16 & 70.6 & 49.3 & 49.4 & 41.5 \\
& SUN+AWQ-d & 16/4 & 11.4 & 7.0 & 43.9 & 38.4 \\
& QSUN    & 16/4  & \textbf{70.3} & \textbf{48.1} & \textbf{47.6} & \textbf{42.7} \\
\midrule
\multirow{3}{*}{\textbf{Qwen3-8B-Base}}
& Full-FT & 16/16 & 85.8 & 65.7 & 83.5 & 74.3 \\ 
& Full-FT+AWQ & 4/4   & 83.6 & 64.0 & 79.3 & 72.6 \\ \cdashline{2-7}\noalign{\vskip 0.35ex}
& SUN & 16/16 & 85.5 & 64.5 & 84.1 & 76.2 \\
& SUN+AWQ-d & 16/4 & 79.8 & 60.4 & 82.3 & 75.6 \\
& QSUN    & 16/4  & \textbf{85.1} & \textbf{64.9} & \textbf{82.3} & \textbf{75.0} \\
\bottomrule
\end{tabular}%
\end{table*}


\clearpage
\section{Performance of QSUN in Higher Concurrency and ISL}
\label{app: qsun performance}

We evaluate the latency and throughput characteristics of full-precision and quantized models under varying concurrency levels.
Table~\ref{tab:QSUN-latency-throughput} reports the time-to-first-token (TTFT), time per output token (TPOT), and overall throughput (TPUT) for different models, methods, and input/output sequence length (ISL/OSL) configurations.

Across all settings, full-precision fine-tuning (Full-FT) exhibits stable latency behavior but limited scalability as concurrency increases. In contrast, quantized methods significantly improve throughput, especially under higher concurrency. AWQ consistently achieves the higher throughput, but at the cost of increased TTFT.

QSUN strikes a favorable balance between latency and throughput. By applying quantization only to the decode phase, QSUN preserves TTFT comparable to Full-FT while achieving substantial throughput gains. For example, under ISL=1024 and OSL=1024, QSUN improves throughput by more than $1.5\times$ at concurrency 16 compared to Full-FT, while maintaining nearly identical TTFT. This trend is consistent across larger ISL settings, demonstrating that QSUN effectively decouples throughput scaling from prefill latency.


Overall, these results highlight that decode-only quantization provides a practical and effective trade-off: it delivers most of the throughput benefits of quantization while retaining the prefill latency characteristics of full-precision models.

\begin{table*}[t]
\centering
\caption{Latency and throughput comparison of Full-FT, AWQ, and QSUN on LLaMA3.1-8B and Qwen3-8B-Base under different concurrency levels (1/4/16). We report time-to-first-token (TTFT, ms), time-per-output-token (TPOT, ms), and throughput (TPUT, tok/s). The \# Bits column denotes the precision of weights (in bits) used in the prefill/decode phases (prefill/decode). ISL/OSL means input sequence length and output sequence length, respectively.}
\label{tab:QSUN-latency-throughput}
\resizebox{\columnwidth}{!}{
\begin{tabular}{c | c  c | ccc ccc ccc}
\toprule
\multirow{2}{*}{\textbf{Model}} 
& \multirow{2}{*}{\textbf{Method}} 
& \multirow{2}{*}{\textbf{\# Bits}} 
& \multicolumn{3}{c}{\textbf{Concurrency = 1}} 
& \multicolumn{3}{c}{\textbf{Concurrency = 4}} 
& \multicolumn{3}{c}{\textbf{Concurrency = 16}} \\
\cmidrule(lr){4-6} \cmidrule(lr){7-9} \cmidrule(lr){10-12}
& & 
& \textbf{TTFT} & \textbf{TPOT} & \textbf{TPUT} 
& \textbf{TTFT} & \textbf{TPOT} & \textbf{TPUT} 
& \textbf{TTFT} & \textbf{TPOT} & \textbf{TPUT} \\
\midrule
\multicolumn{12}{c}{\textbf{ISL = 1024 / OSL = 1024}} \\
\midrule
\multirow{3}{*}{LLaMA3.1-8B}
& Full-FT & 16 / 16 & 99.1  & 13.8 & 73.3  & 179.8 & 15.1 & 265.3 & 140.3 & 17.7 & 908.1 \\
& +AWQ     & 4 / 4   & 118.7 & 7.6  & 130.8 & 230.6 & 8.4  & 463.7 & 197.5 & 11.0 & 1434.4 \\\cdashline{2-12}\noalign{\vskip 0.35ex}
& QSUN    & 16 / 4  & 98.9  & 7.6  & 130.2 & 178.7 & 8.6  & 456.2 & 141.4 & 11.3 & 1400.9 \\
\midrule
\multirow{3}{*}{Qwen3-8B-Base}
& Full-FT & 16 / 16 & 96.4  & 12.2 & 81.3  & 179.5 & 13.0 & 304.5 & 147.1 & 15.2 & 1040.2 \\
& +AWQ     & 4 / 4   & 115.8 & 6.6  & 148.9 & 233.8 & 7.5  & 516.4 & 183.8 & 9.9  & 1580.3 \\ \cdashline{2-12}\noalign{\vskip 0.35ex}
& QSUN    & 16 / 4  & 97.1  & 6.6  & 147.9 & 179.2 & 7.6  & 510.2 & 144.3 & 9.9  & 1580.2 \\
\midrule
\multicolumn{12}{c}{\textbf{ISL = 2048 / OSL = 1024}} \\
\midrule
\multirow{3}{*}{LLaMA3.1-8B}
& Full-FT & 16 / 16 & 176.5 & 15.1 & 66.3  & 377.0 & 15.6 & 253.9 & 231.1 & 19.5 & 824.0 \\
& +AWQ     & 4 / 4   & 210.5 & 8.0  & 122.0 & 470.3 & 9.2  & 412.9 & 260.4 & 13.2 & 1195.0 \\ \cdashline{2-12}\noalign{\vskip 0.35ex}
& QSUN    & 16 / 4  & 175.6 & 7.9  & 124.3 & 376.3 & 9.3  & 405.5 & 224.3 & 13.2 & 1192.1 \\
\midrule
\multirow{3}{*}{Qwen3-8B-Base}
& Full-FT & 16 / 16 & 166.0 & 12.4 & 79.5  & 381.8 & 13.5 & 288.3 & 221.3 & 16.9 & 934.6 \\
& +AWQ     & 4 / 4   & 200.4 & 6.9  & 141.1 & 492.6 & 8.0  & 468.9 & 258.9 & 11.5 & 1356.0 \\ \cdashline{2-12}\noalign{\vskip 0.35ex}
& QSUN    & 16 / 4  & 166.3 & 6.8  & 142.6 & 384.0 & 8.0  & 472.3 & 220.7 & 11.5 & 1360.5 \\
\midrule
\multicolumn{12}{c}{\textbf{ISL = 4096 / OSL = 1024}} \\
\midrule
\multirow{3}{*}{LLaMA3.1-8B}
& Full-FT & 16 / 16 & 317.3 & 14.9 & 66.9  & 638.2 & 16.9 & 230.0 & 643.6 & 24.6 & 641.1 \\
& +AWQ     & 4 / 4   & 384.2 & 7.9  & 121.6 & 861.6 & 10.1 & 363.2 & 730.8 & 17.7 & 871.2 \\ \cdashline{2-12}\noalign{\vskip 0.35ex}
& QSUN    & 16 / 4  & 317.4 & 7.9  & 121.2 & 630.4 & 10.1 & 362.3 & 650.0 & 17.8 & 876.3 \\
\midrule
\multirow{3}{*}{Qwen3-8B-Base}
& Full-FT & 16 / 16 & 327.7 & 12.5 & 77.7  & 686.7 & 14.4 & 265.5 & 768.1 & 21.8 & 708.7 \\
& +AWQ     & 4 / 4   & 398.6 & 6.9  & 136.1 & 897.4 & 8.8  & 411.5 & 811.4 & 16.2 & 938.1 \\ \cdashline{2-12}\noalign{\vskip 0.35ex}
& QSUN    & 16 / 4  & 328.0 & 6.9  & 136.6 & 684.9 & 8.9  & 409.9 & 755.6 & 16.3 & 936.0 \\
\bottomrule
\end{tabular}
}
\end{table*}

\end{document}